\def\year{2021}\relax
\documentclass[letterpaper]{article} % DO NOT CHANGE THIS
\usepackage{aaai21}  % DO NOT CHANGE THIS
\usepackage{times}  % DO NOT CHANGE THIS
\usepackage{helvet} % DO NOT CHANGE THIS
\usepackage{courier}  % DO NOT CHANGE THIS
\usepackage[hyphens]{url}  % DO NOT CHANGE THIS
\usepackage{graphicx} % DO NOT CHANGE THIS
\usepackage{natbib}  % DO NOT CHANGE THIS AND DO NOT ADD ANY OPTIONS TO IT
\usepackage{caption} % DO NOT CHANGE THIS AND DO NOT ADD ANY OPTIONS TO IT
\usepackage{url}
\usepackage{amsfonts}
\usepackage{amsmath}
\usepackage{booktabs}

\usepackage{multirow}
\usepackage{subfigure}
\usepackage{amssymb}
\usepackage{bbding}
\usepackage[switch]{lineno}
\usepackage{bm}
\title{Dynamic Anchor Learning for Arbitrary-Oriented Object Detection}
\author{
    Qi Ming, Zhiqiang Zhou\thanks{Corresponding author}, Lingjuan Miao, Hongwei Zhang, Linhao Li
	\\
}

\affiliations{
	School of Automation, Beijing Institute of Technology, China\\
	chaser.ming@gmail.com, \{zhzhzhou, miaolingjuan\}@bit.edu.cn, \\zhanghw.hongwei@gmail.com, lilinhao@bit.edu.cn \\

}
\nocopyright
\begin{document}
%\linenumbers
\maketitle

\begin{abstract}

Arbitrary-oriented objects widely appear in natural scenes, aerial photographs, remote sensing images, etc., and thus arbitrary-oriented object detection has received considerable attention. Many current rotation detectors use plenty of anchors with different orientations to achieve spatial alignment with ground truth boxes. Intersection-over-Union (IoU) is then applied to sample the positive and negative candidates for training. However, we observe that the selected positive anchors cannot always ensure accurate detections after regression, while some negative samples can achieve accurate localization. It indicates that the quality assessment of anchors through IoU is not appropriate, and this further leads to inconsistency between classification confidence and localization accuracy. In this paper, we propose a dynamic anchor learning (DAL) method, which utilizes the newly defined matching degree to comprehensively evaluate the localization potential of the anchors and carries out a more efficient label assignment process. In this way, the detector can dynamically select high-quality anchors to achieve accurate object detection, and the divergence between classification and regression will be alleviated. With the newly introduced DAL, we can achieve superior detection performance for arbitrary-oriented objects with only a few horizontal preset anchors. Experimental results on three remote sensing datasets HRSC2016, DOTA, UCAS-AOD as well as a scene text dataset ICDAR 2015 show that our method achieves substantial improvement compared with the baseline model. Besides, our approach is also universal for object detection using horizontal bound box. The code and models are available at https://github.com/ming71/DAL.
\end{abstract}

\section{Introduction}

Object detection is one of the most fundamental and challenging problem in computer vision. In recent years, with the development of deep convolutional neural networks (CNN) , tremendous successes have been achieved on object detection \cite{ren2015faster, dai2016r, redmon2016you, liu2016ssd}. Most detection frameworks utilize preset horizontal anchors to achieve spatial alignment with ground-truth(GT) box. Positive and negative samples are then selected through a specific strategy during training phase, which is called label assignment. 

Since objects in the real scene tend to appear in diverse orientations, the issue of oriented object detection has gradually received considerable attention. There are many approaches have achieved oriented object detection by introducing the extra orientation prediction and preset rotated anchors \cite{ma2018arbitrary, liao2018textboxes++}. These detectors often follow the same label assignment strategy as general object detection frameworks. For simplicity, we call the IoU between GT box and anchors as input IoU, and the IoU between GT box and regression box as output IoU. The selected positives tend to obtain higher output IoU compared with negatives, because their better spatial alignment with GT provides sufficient semantic knowledge, which is conducive to accurate classification and regression.

\begin{figure}[t]
	\subfigure[]{
		\includegraphics[width=0.232\textwidth]{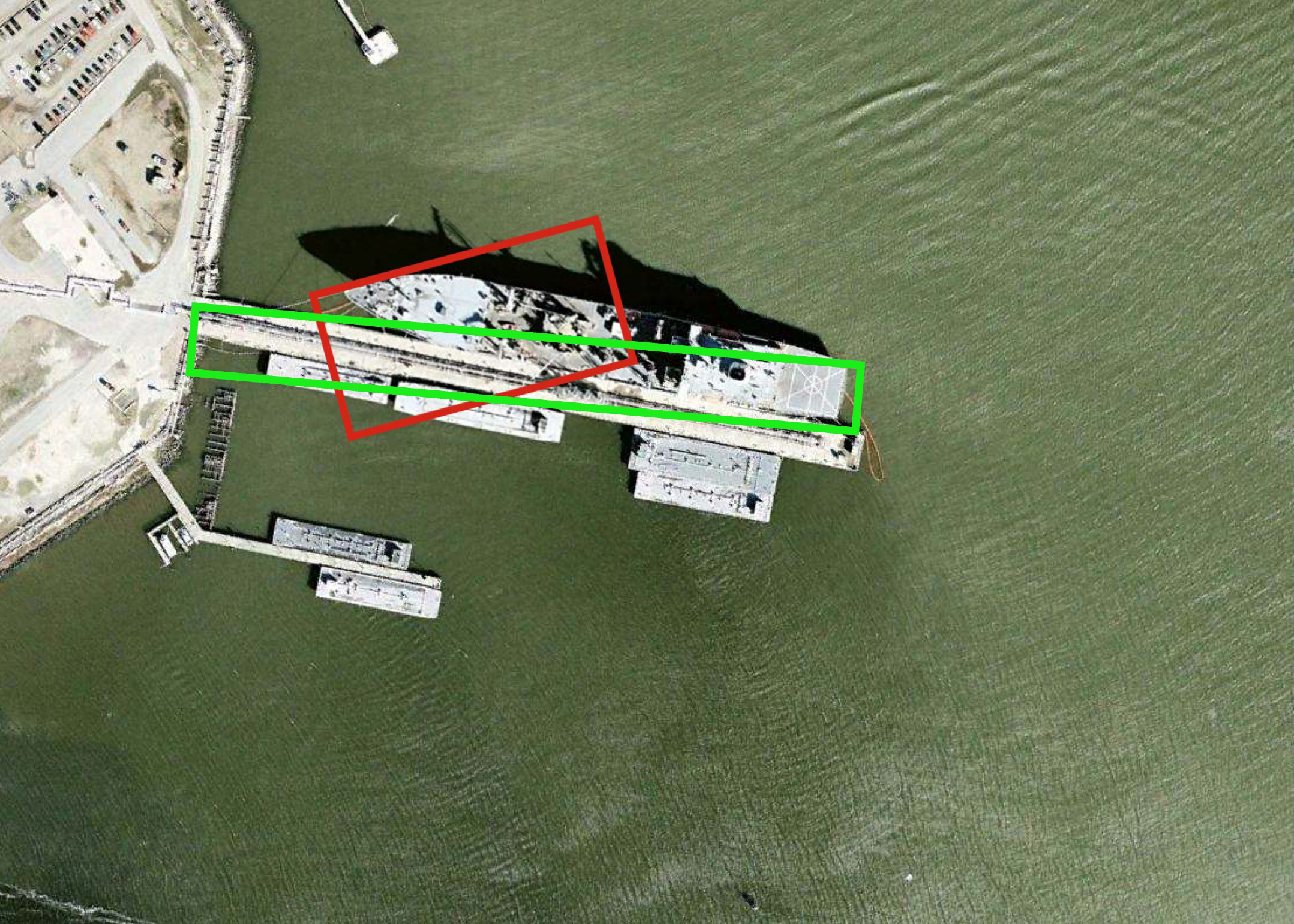}}\hspace{-1mm}
	\subfigure[]{
		\includegraphics[width=0.232\textwidth]{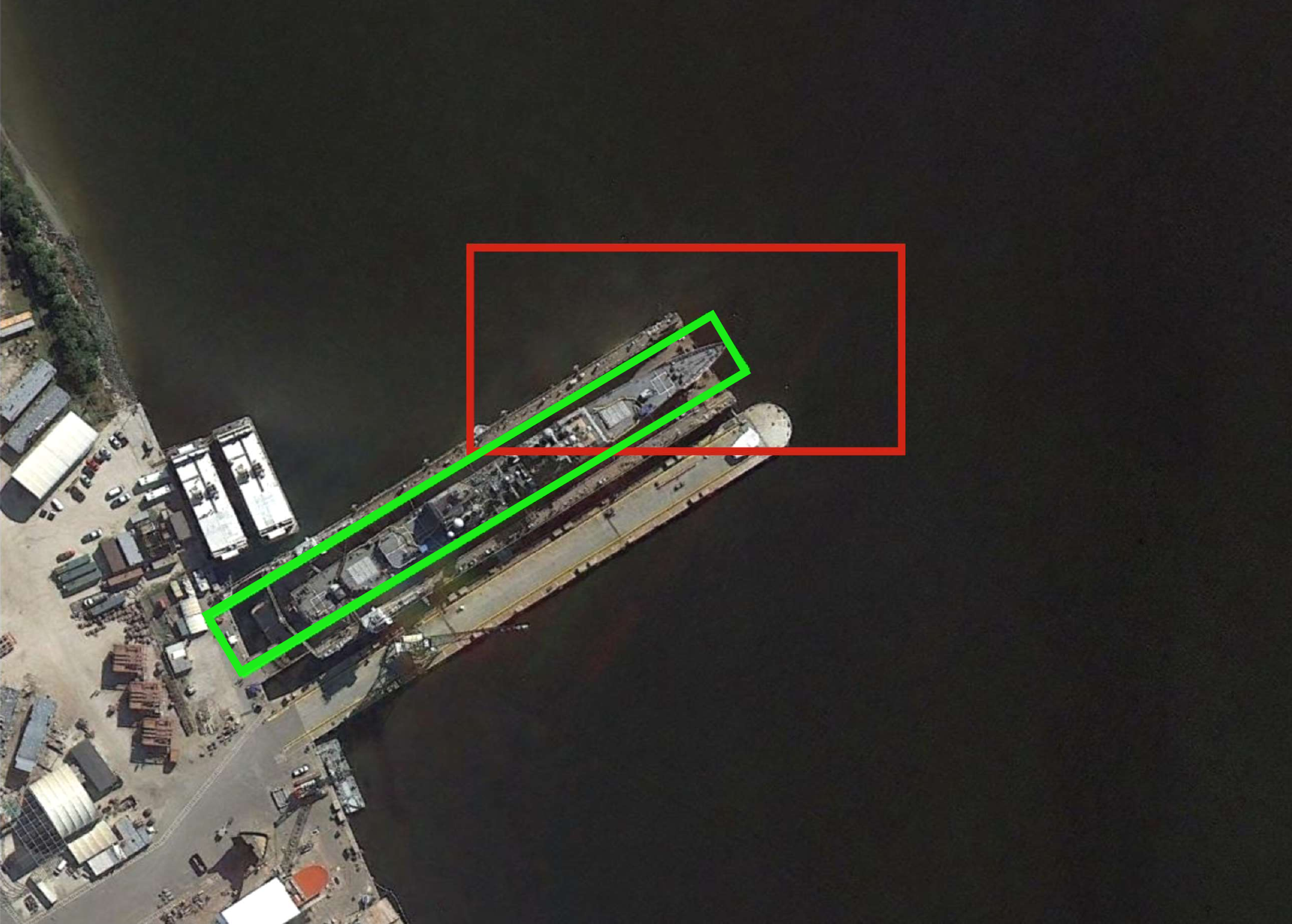}}
	\caption{Predefined anchor (red) and its regression box (green). (a) shows that anchors with a high input IoU cannot guarantee perfect detection. (b) reveals that the anchor that is poorly spatially aligned with the GT box still has the potential to localize object accurately.}
	\label{fig1}
\end{figure}
\begin{figure}[t]
	\subfigure[]{
		\label{Fig2a}
		\includegraphics[width=0.22\textwidth]{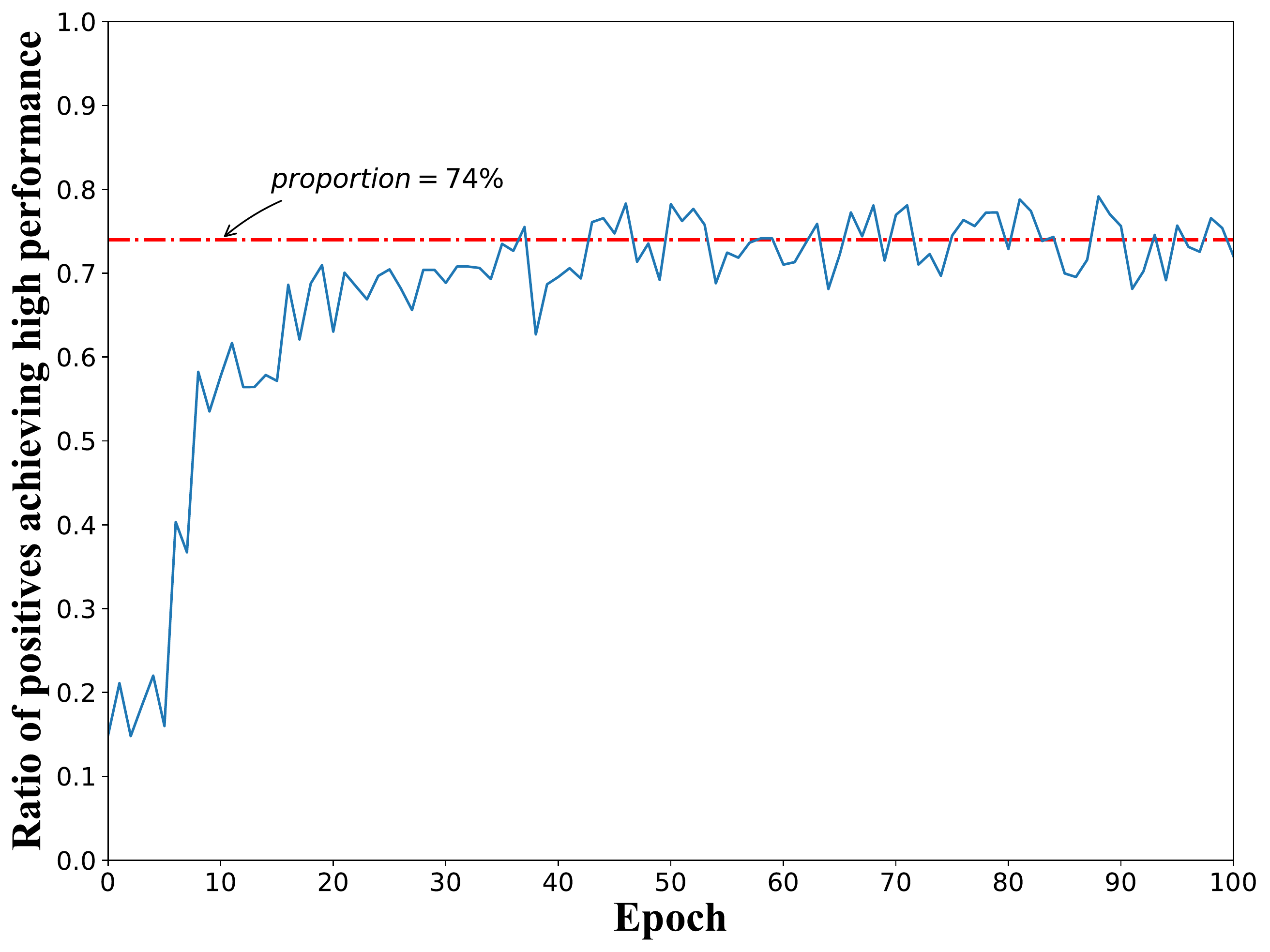}}\hspace{-1mm}
	\subfigure[]{
		\label{Fig2b}
		\includegraphics[width=0.22\textwidth]{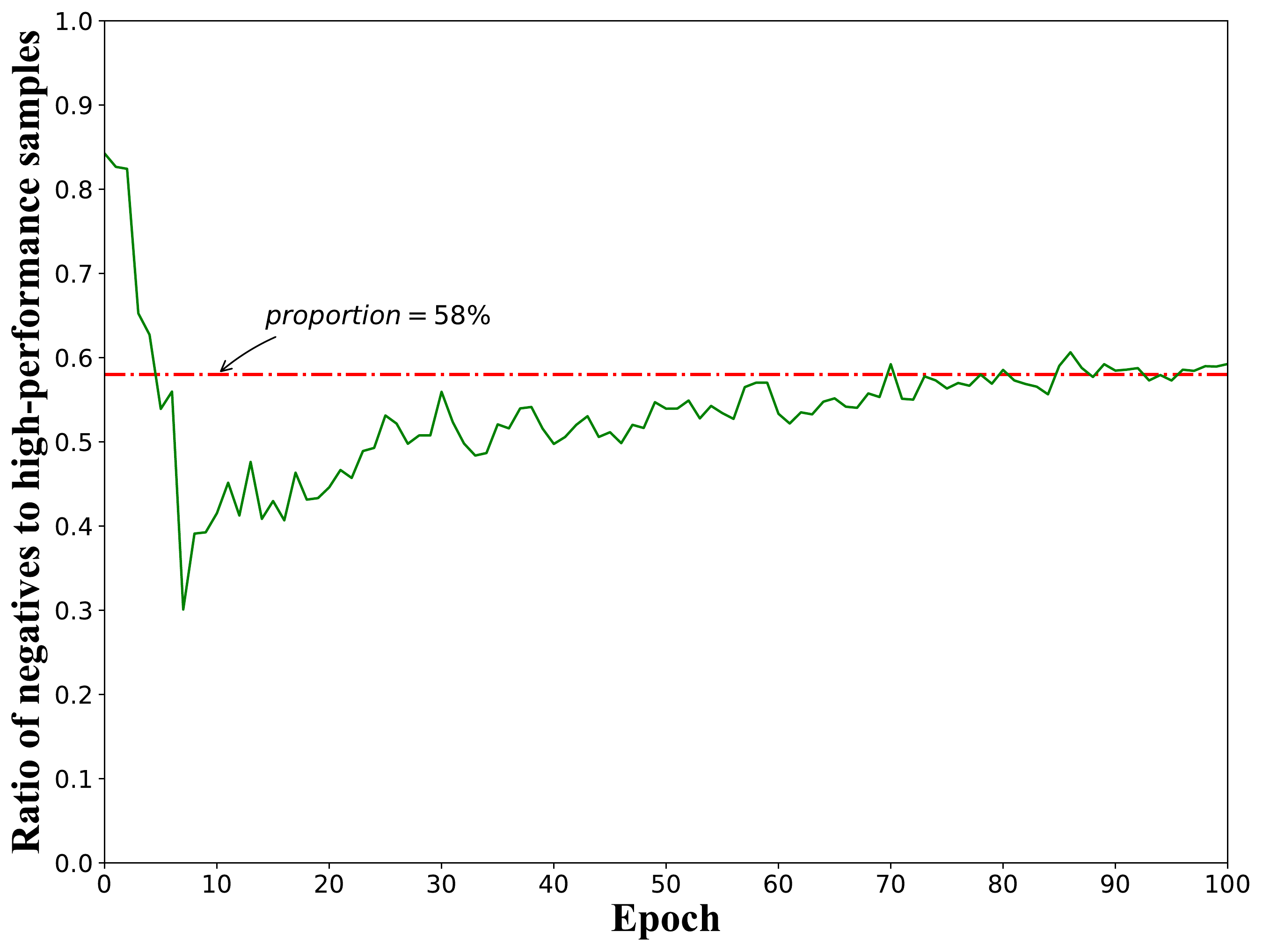}}
	\subfigure[]{
		\label{Fig2c}
		\includegraphics[width=0.22\textwidth]{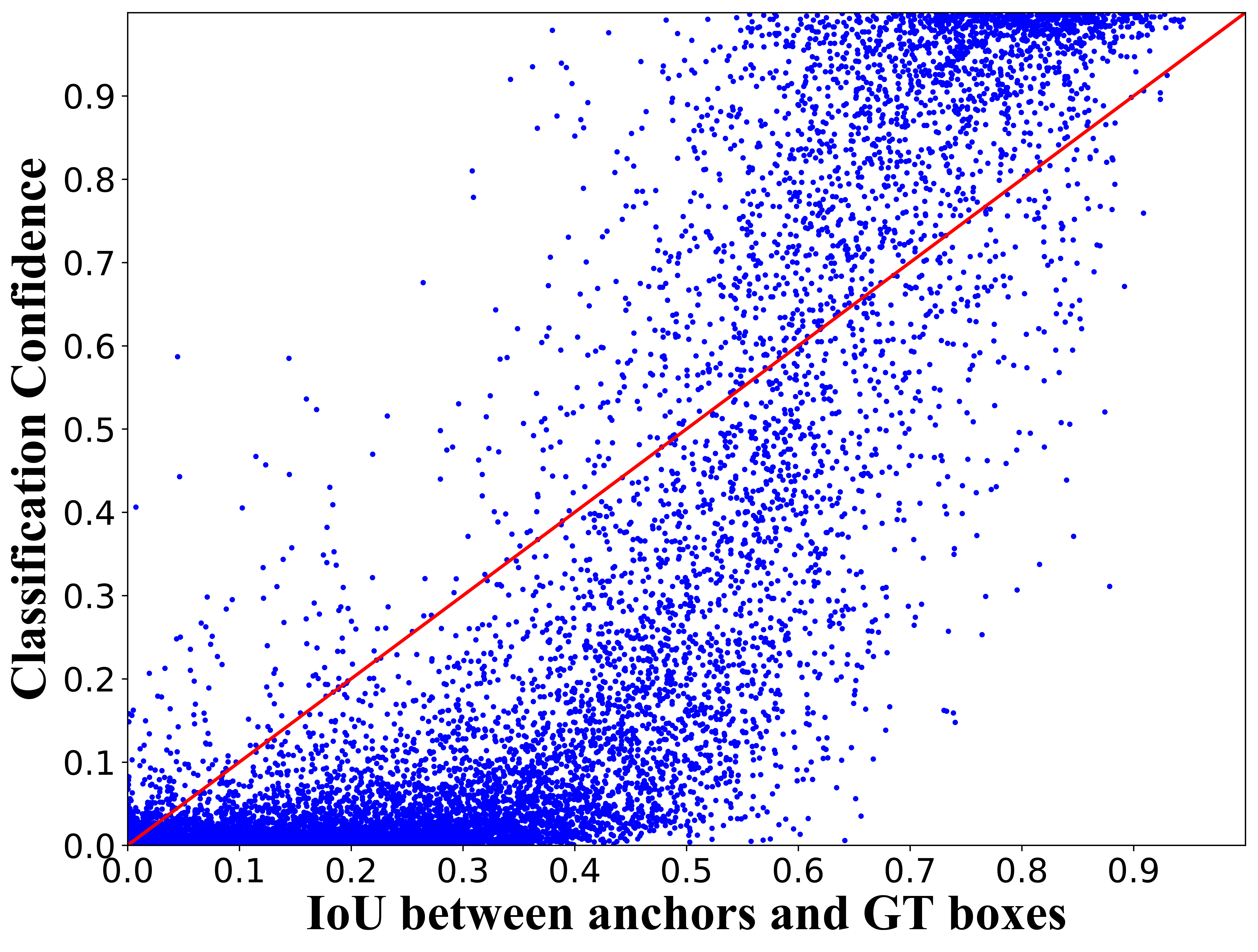}}\hspace{3mm}
	\subfigure[]{
		\label{Fig2d}
		\includegraphics[width=0.22\textwidth]{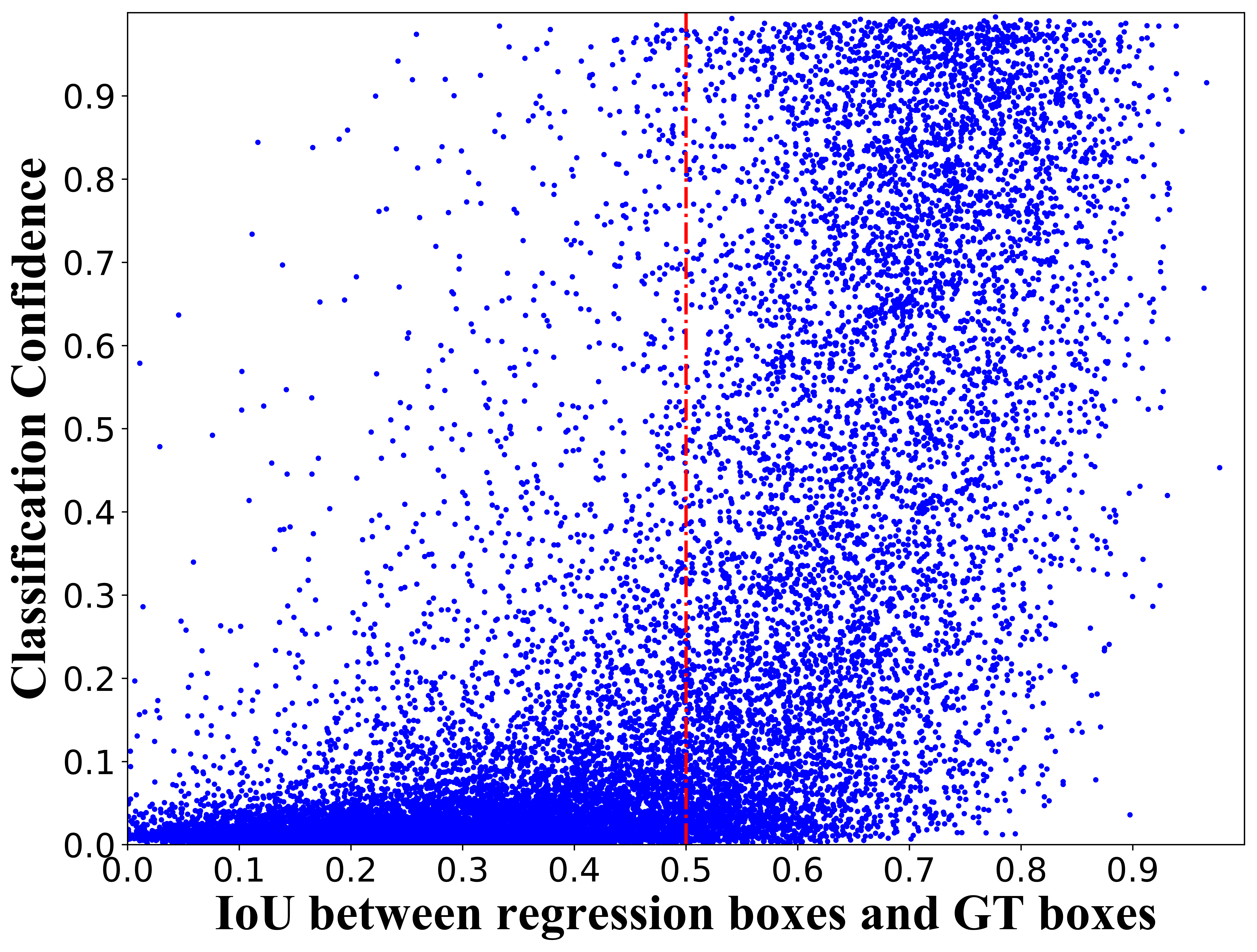}}
	\caption{Analysis of the classification and regression capabilities of anchors that use input IoU for label assignment. (a) Only 74\% of the positive sample anchors can localize GT well after regression (with output IoU higher than 0.5), which illustrates that many false positive samples are introduced. (b) Only 42\% of the high-quality detections (output IoU is higher than 0.5) come from matched anchors, which implies that quite a lot of negative anchors (58\% in this example) have the potential to achieve accurate localization. (c) Current label assignment leads to a positive correlation between the classification confidence and the input IoU. (d) High-performance detection results exhibit a weak correlation between the localization ability and classification confidence, which is not conducive to selecting accurate detection results through classification score during inference.}
	\label{fig2}
\end{figure}

However, we observe that the localization performance of the assigned samples is not consistent with the assumption mentioned above. As shown in Figure \ref{fig1}, the division of positive and negative anchors seems not always related to the detection performance. Furthermore, we have counted the distribution of the anchor localization performance for all candidates to explore whether this phenomenon is universal. As illustrated in Figure \ref{Fig2a},  a considerable percentage (26\%) of  positive anchors are poorly aligned with GT after regression, revealing that the positive anchors cannot ensure accurate localization. Besides, more than half of the candidates that achieve high-quality predictions are regressed from negatives (see Figure \ref{Fig2b}) which  implies that a considerable number of negatives with high localization potential have not been effectively used. In summary, we conclude that the localization performance does not entirely depend on the spatial alignment between the anchors and GT. 

Besides, inconsistent localization performance before and after anchor regression further lead to inconsistency between classification and localization, which has been discussed in previous work \cite{jiang2018acquisition, kong2019consistent, he2019bounding}. As shown in Figure \ref{Fig2c}, anchor matching strategy based on input IoU induces a positive correlation between the classification confidence and input IoU. However, as discussed above, the input IoU is not entirely equivalent to the localization performance. Therefore, we cannot distinguish the localization performance of the detection results based on the classification score. The results in Figure \ref{Fig2d} also confirm this viewpoint: a large number of regression boxes with high output IoU  are misjudged as background.

To solve the problems, we propose a Dynamic Anchor Learning(DAL) method for better label assignment and further improve the detection performance. Firstly, a simple yet effective standard named \textit{matching degree} is designed to assess the localization potential of anchors, which comprehensively considers the prior information of spatial alignment, the localization ability and the regression uncertainty. After that, we adopt matching degree for training sample selection, which helps to eliminate false-positive samples and dynamically mine potential high-quality candidates, as well as suppress the disturbance caused by regression uncertainty. Next, we propose a matching-sensitive loss function to further alleviate the inconsistency between classification and regression, making the classifier more discriminative for proposals with high localization performance, and ultimately achieving high-quality detection.

Extensive experiments on public datasets, including remote sensing datasets HRSC2016, DOTA, UCAS-AOD, and scene text dataset ICDAR 2015, show that our method can achieve stable and substantial improvements for arbitrary-oriented object detections. Integrated with our approach, even the vanilla one-stage detector can be competitive with state-of-the-art methods on several datasets. In addition, experiments on ICDAR 2013 and NWPU VHR-10 prove that our approach is also universal for object detection using horizontal box. The proposed DAL approach is general and can be easily integrated into existing object detection pipeline without increasing the computational cost of inference. 

Our contributions are summarized as follows:

\begin{itemize}
	\item We observe that the label assignment based on IoU between anchor and GT box leads to suboptimal localization ability assessment, and further brings inconsistent classification and regression performance.
	\item The matching degree is introduced to measure the localization potential of anchors. A novel label assignment method based on this metric is proposed to achieve high-quality detection.
	\item The matching-sensitive loss is proposed to alleviate the problem of the weak correlation between classification and regression, and improves the discrimination ability of high-quality proposals.
\end{itemize} 

\section{Related Work}
\subsection{Arbitrary-Oriented Object Detection}
The current mainstream detectors can be divided into two categories: two-stage detector \cite{ren2015faster, dai2016r} and one-stage detector \cite{redmon2016you, liu2016ssd}. Existing rotation detectors are mostly built on detectors using horizontal bounding box representation. To localize rotated objects, the preset rotated anchor and additional angle prediction are adopted  in the literature \cite{liu2018arbitrary, ma2018arbitrary, liao2018textboxes++, liu2017rotated}. Nevertheless, due to the variation of the orientation, these detectors are obliged to preset plenty of rotated anchors to make them spatially aligned with GT box. There are also some methods that detect oriented objects only using horizontal anchors. For example, RoI Transformer \cite{ding2019learning} uses horizontal anchors but learns the rotated RoI through spatial transformation, reducing the number of predefined anchors.  R$^3$Det\cite{yang2019r3det} adopts cascade regression and refined box re-encoding module to achieve state-of-the-art performance with horizontal anchors. Although the above approaches have achieved good performance, they cannot make a correct judgment on the quality of anchors, and thus cause improper label assignment which brings adverse impact during the training process.
\subsection{Label Assignment}
Most anchor-based detectors densely preset anchors at each position of feature maps. The massive preset anchors lead to serious imbalance problem, especially for the arbitrary-oriented objects with additional angle setting. The most common solution is to control the ratio of candidates through a specific sampling strategy \cite{shrivastava2016training, pang2019libra}. Besides, Focal Loss \cite{lin2017focal} lowers the weight of easy examples to avoid its overwhelming contribution to loss. The work \cite{li2019gradient} further considers extremely hard samples as outliers, and use gradient harmonizing mechanism to conquer the imbalance problems. We demonstrate that the existence of outliers is universal, and our method can prevent such noise samples from being assigned incorrectly.

Some work have observed problems caused by using input IoU as a standard for label assignment. Dynamic R-CNN \cite{zhang2020dynamic} and ATSS \cite{zhang2020bridging} automatically adjust the IoU threshold to select high-quality positive samples, but they fail to consider whether the IoU itself is credible. The work \cite{li2020learning} points out that binary labels assigned to anchors are noisy, and construct cleanliness score for each anchor to supervise training process. However, it only considers the noise of positive samples, ignoring the potentially powerful localization capabilities of massive negatives.  HAMBox \cite{liu2020hambox} reveals that unmatched anchors can also achieve accurate prediction, and attempts to utilize these samples. Nevertheless, its compensated anchors mined according to output IoU is not reliable; moreover, it does not consider the degradation of matched positives. FreeAnchor \cite{zhang2019freeanchor} formulates object-anchor matching as a maximum likelihood estimation procedure to select the most representative anchors, but its formulation is relatively complicated.

\section{Proposed Method}

\subsection{Rotation Detector Built on RetinaNet}
The real-time inference is essential for arbitrary-oriented object detection in many scenarios. Hence we use the one-stage detector RetinaNet \cite{lin2017focal} as the baseline model. It utilizes ResNet-50 as backbone, in which the architecture similar to FPN \cite{lin2017feature} is adopted to construct a multi-scale feature pyramid. Predefined horizontal anchors are set on the features of each level $P_3$, $P_4$, $P_5$, $P_6$, $P_7$. Note that rotation anchor is not used here, because it is inefficient and unnecessary, and we will further prove this point in the next sections. 
Since the extra angle parameter is introduced, the oriented  box is represented in the format of $(x, y, w, h, \theta)$. For bounding box regression, we have:

\begin{equation} 
\begin{aligned}
t_{x}&=\left(x-x_{a}\right) / w_{a}, \quad t_{y}=\left(y-y_{a}\right) / h_{a} \\
t_{w}&=\log \left(w / w_{a}\right), \quad \quad t_{h}=\log \left(h / h_{a}\right) \\
t_{\theta}&=\tan \left(\theta-\theta_{a}\right)
\end{aligned}
\end{equation}
where $x, y, w, h, \theta$ denote center coordinates, width, height and angle, respectively. $x$ and $x_a$ are for the predicted box and anchor, respectively (likewise for $y, w, h, \theta$). Given the ground-truth box offsets $\boldsymbol{t}^{*}=\left(t_{x}^{*}, t_{y}^{*}, t_{w}^{*}, t_{h}^{*},  t_{\theta}^{*}\right)$, the multi-task loss is defined as follows:
\begin{equation} 
L=L_{c l s}(p, p^{*})+L_{r e g}\left(\boldsymbol{t}, \boldsymbol{t}^{*}\right)
\end{equation}
in which the value $p$ and the vector $\boldsymbol{t}$ denote predicted classification score and predicted box offsets, respectively. Variable $p^*$ represents the class label for anchors ($p^*=1$ for positive samples and $p^*=0$ for negative samples).

\subsection{Dynamic Anchor Selection}
Some researches \cite{zhang2019freeanchor, song2020revisiting} have reported that the discriminative features required to localize objects are not evenly distributed on GT, especially for objects with a wide variety of orientations and aspect ratios. Therefore, the label assignment strategy based on spatial alignment, i.e., input IoU, leads to the incapability to capture the critical feature demanded for object detection. 

An intuitive approach is to use the feedback of the regression results, that is, the output IoU to represent feature alignment ability and dynamiclly guide the training process. Several attempts \cite{jiang2018acquisition, li2020learning} have been made in this respect. In particular, we tentatively select the training samples based on output IoU and use it as soft-label for classification. However, we found that the model is hard to  converge because of the following two issues:

\begin{itemize}
	\item Anchors with high input IoU but low output IoU are not always negative samples, which may be caused by not sufficient training.
	\item The unmatched low-quality anchors that accidentally achieve accurate localization performance tend to be misjudged as positive samples.
\end{itemize}

The above analysis shows that regression uncertainty interferes with the credibility of the output IoU for feature alignment. Regression uncertainty has been widely discussed in many previous work \cite{feng2018towards, choi2019gaussian, kendall2017uncertainties, choi2018uncertainty}, which represents the instability and irrelevance in the regression process. We discovered in the experiment that it also misleads the label assignment. Specifically, high-quality samples cannot be effectively utilized, and the selected false-positive samples would cause the unstable training. Unfortunately, neither the input IoU nor the output IoU used for label assignment can avoid the interference caused by the regression uncertainty.

Based on the above observations, we introduce the concept of \textit{matching degree} (MD), which  utilizes the prior information of spatial matching, feature alignment ability and regression uncertainty of the anchor to measure the localization capacity, which is defined as follows: 
\begin{equation} \label{eq3}
m d=\alpha \cdot s a+(1-\alpha) \cdot f a-u^{\gamma}
\end{equation}
where $sa$ denotes a priori of spatial alignment, whose value is equivalent to input IoU. $fa$ represents the feature alignment capability calculated by IoU between GT box and regression box. $\alpha$ and $\gamma$ are hyperparameters used to weight the influence of different items. $u$ is a penalty term, which denotes the regression uncertainty during training. It is obtained via the IoU variation before and after regression:
\begin{equation} 
u=|sa-fa|
\end{equation}

The suppression of interference  during regression is vital for high-quality anchor sampling and stable training. Variation of IoU before and after regression represents the probability of incorrect anchor assessment. Note that our construction of the penalty term for regression uncertainty is very simple, and since detection performance is not sensitive to the form of $u$ , the naive, intuitive yet effective form is employed.

With the newly defined matching degree, we  conduct dynamic anchor selection for superior label assignment. In the training phase, we first calculate matching degree between the GT box and anchors, and then the anchors with matching degree higher than a certain threshold (set to 0.6 in our experiments) are selected as positives, while the rest are negatives. After that, for GT that do not match any anchor, the anchor with the highest matching degree will be compensated as positive candidate. To achieve more stable training,  we gradually adjust the impact of the input IoU during training. The specific adjustment schedule is as follows:
\begin{equation} 
\begin{aligned}
\alpha(t)\!=\!\left\{\!
\begin{array}{lcl}
1,       & & {t      <      0.1}\\
5(\alpha_0-1)\cdot \!t\!+\!1.5\!-\! 0.5 \!\cdot \!\alpha_0,       & & {0.1 \!\leq\! t\! < \!0.3}\\
\alpha_0,          & & {t\geq0.3}\\
\end{array} \right. 
\end{aligned}
\end{equation} 
where $t= \frac{iters}{Max\_Iteration}$, $Max\_Iteration$ is the total number of iterations, and $\alpha_0$ is the final weighting factor that appears in Eq. (\ref{eq3}).

\begin{figure}[t]
	\subfigure[Without MSL]{
		\label{fig3a}
		\includegraphics[width=0.23\textwidth]{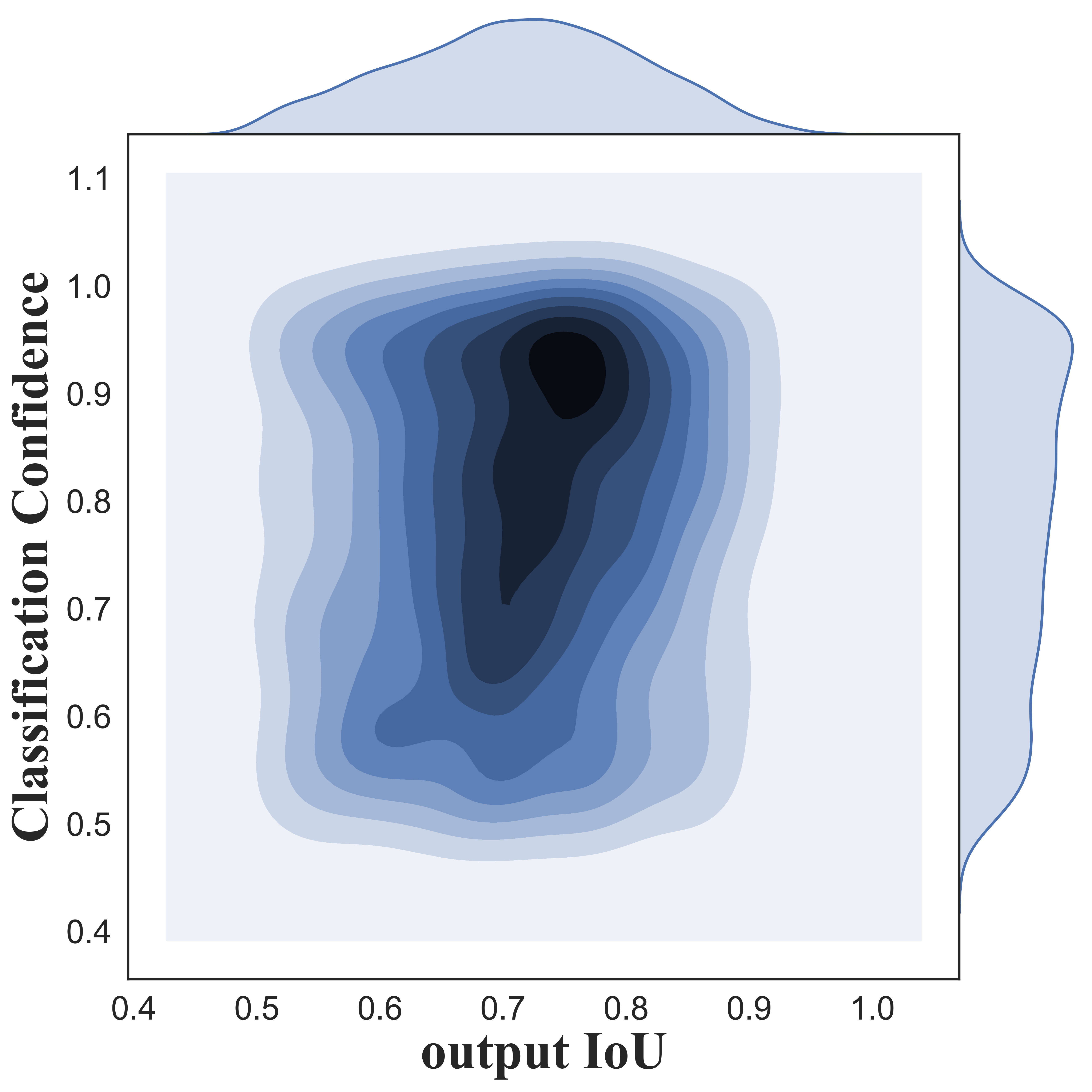}}\hspace{-1mm}
	\subfigure[With MSL]{
		\label{fig3b}
		\includegraphics[width=0.23\textwidth]{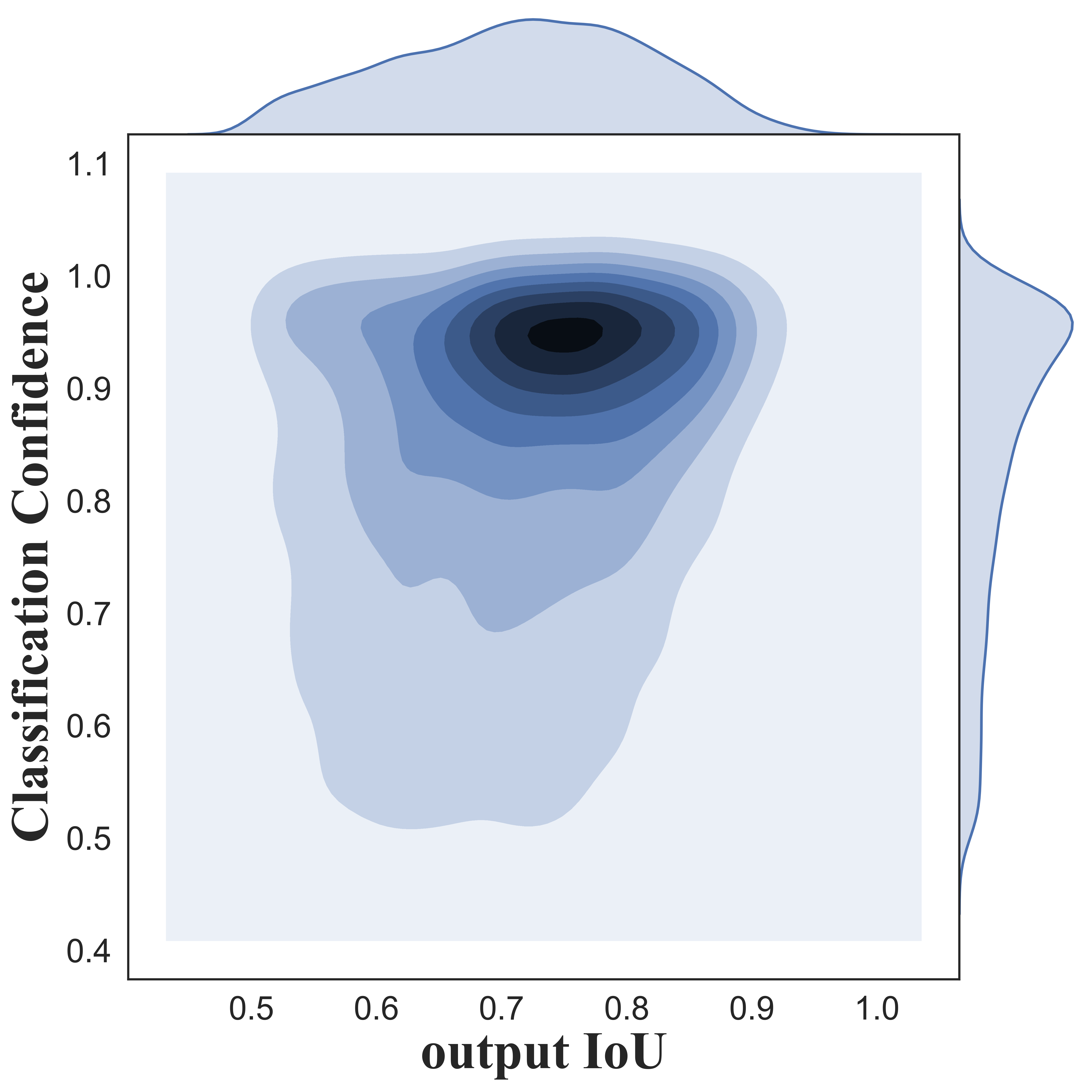}}
	\caption{ The correlation between the output IoU and classification score with and without MSL.}
	\label{fig3}
\end{figure}

\subsection{Matching-Sensitive Loss }
To further enhance the correlation between classification and regression to achieve high-quality arbitrary-oriented detection, we integrate the matching degree into the training process and propose the matching-sensitive loss function (MSL). The classification loss is defined as:
\begin{equation} 
L_{cls}=\frac{1}{N}\sum_{i \in \psi} FL\left(p_{i}, p_{i}^{*}\right) + \frac{1}{N_{p}}\sum_{j \in \psi_p} w_j \cdot FL\left(p_{j}, p_{j}^{*}\right)
\end{equation}
where $\psi$ and  $\psi_p$ respectively represent all anchors and the positive samples selected by the matching degree threshold. $N$ and $N_p$  denote the total number of all anchors and positive anchors, respectively. $FL(\cdot)$  is focal loss defined in RetinaNet \cite{lin2017focal}. $w_j$ indicates the matching compensation factor, which is used to distinguish positive samples of different localization potential. For each ground-truth $g$, we first calculate its matching degree with all anchors as $\bm{md}$. Then positive candidates can be selected according to a certain threshold, matching degree of positives is represented as $ \bm{md}_{pos}$, where $ \bm{md}_{pos} \subseteq \bm{md} $. Supposing that the maximal matching degree for $g$ is $md_{\text{max}}$, the compensation value is denoted as $\Delta md$, we have:
\begin{equation} 
\Delta md = 1- md_{\text{max}}.
\end{equation}

After that, $\Delta md$  is added to the matching degree of all positives to form the matching compensation factor:
\begin{equation} 
\bm{w} =  \bm{md}_{pos}+\Delta md.
\end{equation}

With the well-designed matching compensation factor, the detector treats positive samples of different localization capability distinctively. In particular, more attention will be paid to candidates with high localization potential for classifier. Therefore, high-quality predictions can be taken through classification score, which helps to alleviate the inconsistency between classification and regression.

Since matching degree measures the localization ability of anchors, and thus it can be further used to promote high-quality localization. We formulate the matching-sensitive regression loss as follows:
\begin{equation} 
L_{reg}=\frac{1}{N_p}\sum_{j \in \psi_p} w_j \cdot L_{smooth_{L_1}}\left(\bm{t_{j}}, \bm{t_{j}^{*}}\right)
\end{equation}
where $L_{smooth_{L_1}}$ denotes the smooth-$L_1$ loss for regression. Matching compensation factor $\bm{w}$ is embedded into regression loss to avoid the loss contribution of high-quality positives being submerged in the dominant loss of samples with poor spatial alignment with GT boxes.
It can be seen from Figure \ref{fig3a} that the correlation between the classification score and the localization ability of the regression box is not strong enough, which causes the prediction results selected by the classification confidence to be sometimes unreliable. After training with a matching-sensitive loss, as shown in Figure \ref{fig3b}, a higher classification score accurately characterizes the better localization performance represented by the output IoU, which verifies the effectiveness of the proposed method.
\section{Experiments}
\subsection{Datasets}
We conduct experiments on the remote sensing datasets HRSC2016, DOTA, UCAS-AOD and a scene text dataset ICDAR 2015. The ground-truth boxes in the images are all oriented bounding boxes. The HRSC2016 \cite{lb2017high} is a challenging remote sensing ship detection dataset, which contains 1061 pictures. The entire dataset is divided into training set, validation set and test set, including 436, 541 and 444 images, respectively. DOTA \cite{xia2018dota}  is the largest public dataset for object detection in remote sensing imagery with oriented bounding box annotations. It contains 2806 aerial images with 188,282 annotated instances, there are 15 categories in total. Note that images in DOTA are too large, we crop images into 800$\times$800 patches with the stride set to 200. UCAS-AOD \cite{zhu2015orientation} is an aerial aircraft and car detection dataset, which contains 1510 images. We  randomly divide it into training set, validation set and test set as 5:2:3. The ICDAR 2015 dataset is used for Incidental Scene Text Challenge 4 of the ICDAR Robust Text Detection Challenge \cite{karatzas2015icdar}. It contains 1500 images, including 1000 training images and 500 test images.          
\begin{table}[t]
	\centering
	\small
	\setlength{\tabcolsep}{2.5mm}{
	\begin{tabular}{{c|cccc}}
		\toprule
		& \multicolumn{4}{c}{Different Variants} \\ 
		\midrule
		with Input IoU &\checkmark &\checkmark & \checkmark&\checkmark \\ 
		with Output IoU&          &\checkmark &\checkmark&\checkmark\\ 
		Uncertainty Supression&          &          &\checkmark&\checkmark\\ 
		Matching Sensitive Loss&          &          &         &\checkmark\\ 
		\midrule
		$\text{AP}_{50}$&     80.8     &    78.9      &    85.9     &   $\textbf{88.6}$      \\ 
		$\text{AP}_{75}$&     52.4     &      50.4    &   57.7      &     $\textbf{67.6}$   \\ 
		\bottomrule
	\end{tabular}}
\caption{Effects of each component in our method on HRSC2016 dataset.}
	\label{table1}
\end{table}
\begin{table}[t]
	\centering
	\begin{tabular}{c|cc|c|cc|c|cc}
		\toprule
		$\gamma$     & $\alpha$   & $\text{mAP}$& $\gamma$   & $\alpha$   & $\text{mAP}$ & 	$\gamma$   &  $\alpha$& $\text{mAP}$   \\
		\midrule
		\multirow{5}{*}{3} & 0.2 &84.1& \multirow{5}{*}{4} &0.2&88.1 & \multirow{5}{*}{5} & 0.2 & 87.3 \\
		& 0.3 & 88.3 &                    & 0.3 & 88.2 &                    & 0.3 & $\textbf{88.6}$ \\ 
& 0.5 & 86.2 &                    & 0.5 & 85.5 &                    & 0.5 & 88.4 \\ 
& 0.7 & 84.1 &                    & 0.7 & 77.9 &                    & 0.7 & 88.1 \\ 
& 0.9 & 70.1 &                    & 0.9 & 75.5 &                    & 0.9 & 83.5 \\
\bottomrule
	\end{tabular}
\caption{Analysis of different hyperparameters on HRSC2016 dataset.}
\label{table2}
\end{table}
\begin{table*}[t]
	\centering
	\setlength{\tabcolsep}{3mm}{
	\begin{tabular}{cccc|c}
		\toprule
		Baseline &\cite{yang2018automatic} &HAMBox\cite{liu2020hambox} & ATSS\cite{zhang2020bridging}&DAL(Ours) \\ 
		\midrule
		80.8 &82.2 &85.4 & 86.1&\textbf{88.6} \\ 
		\bottomrule
	\end{tabular}}
\caption{ Comparisons with other label assignment strategies on HRSC2016.}
\label{table3}
\end{table*}
\begin{table*}[t]
	\scriptsize
	\centering
	\setlength{\tabcolsep}{1.1mm}{
	\begin{tabular}{c|c|ccccccccccccccc|c}
		\toprule
		Methods & Backbone  & PL & BD & BR & GTF & SV & LV & SH & TC & BC & ST & SBF & RA & HA & SP & HC & mAP \\ 
		\midrule
		FR-O\cite{xia2018dota}         & R-101                   & 79.09  &69.12  &17.17  &63.49   &34.20  &37.16  &36.20  &89.19  &69.60  &58.96  &49.40  &52.52  &46.69  &44.80  &46.30  &52.93\\ 
		R-DFPN\cite{yang2018automatic} & R-101                   & 80.92  &65.82  &33.77  &58.94   &55.77  &50.94  &54.78  &90.33  &66.34  &68.66  &48.73  &51.76  &55.10  &51.32  &35.88  &57.94 \\
		$\text{R}^2\text{CNN}$\cite{jiang2017r2cnn} & R-101      & 80.94  &65.67  &35.34  &67.44   &59.92  &50.91  &55.81  &90.67  &66.92  &72.39  &55.06  &52.23  &55.14  &53.35  &48.22  &60.67 \\ 
		RRPN\cite{ma2018arbitrary}	   & R-101                   & 88.52  &71.20  &31.66  &59.30   &51.85  &56.19  &57.25  &90.81  &72.84  &67.38  &56.69  &52.84  &53.08  &51.94  &53.58  &61.01\\ 
		ICN\cite{azimi2018towards} 	   &  R-101                  & 81.36  &74.30  &47.70  &70.32   &64.89  &67.82  &69.98  &90.76  &79.06  &78.20  &53.64  &62.90  &67.02  &64.17  &50.23  &68.16\\
		RoI Trans.\cite{ding2019learning}& R-101                 & 88.64  &78.52  &43.44  &$\textbf{75.92}$   &68.81  &73.68  &83.59  &90.74  &77.27  &81.46  &58.39  &53.54  &62.83  &58.93  &47.67  &69.56\\
		CAD-Net\cite{zhang2019cad}      &  R-101                 & 87.80  &82.40  &49.40  &73.50   &71.10  &63.50  &76.70  &$\textbf{90.90}$  &79.20  &73.30  &48.40  &60.90  &62.00  &67.00  &62.20  &69.90 \\ 
		DRN\cite{pan2020dynamic}        &  H-104                 & 88.91  &80.22  &43.52  &63.35   &73.48  &70.69  &84.94  &90.14  &83.85  &84.11  &50.12  &58.41  &67.62  &68.60  &52.50  &70.70  \\ 
		$\text{O}^2\text{-DNet}$\cite{wei2019oriented} &  H-104  & 89.31  &82.14  &47.33  &61.21   &71.32  &74.03  &78.62  &90.76  &82.23  &81.36  &60.93  &60.17  &58.21  &66.98  &61.03  &71.04\\
		SCRDet\cite{yang2019scrdet}	 &  R-101                  & 89.98  &80.65  &52.09  &68.36   &68.36  &60.32  &72.41  &90.85  &$\textbf{87.94}$  &86.86  &65.02  &66.68  &66.25  &68.24  &65.21  &72.61\\ 
		$\text{R}^3\text{Det}$\cite{yang2019r3det}   &  R-152        & 89.49  &81.17  &50.53  &66.10   &70.92  &78.66  &78.21  &90.81  &85.26  &84.23  &61.81  &63.77  &68.16  &69.83  &67.17  &73.74 \\ 
		CSL\cite{yang2020arbitrary}   &  R-152    &$\textbf{90.25}$  &$\textbf{85.53}$ &54.64  &75.31   &70.44  &73.51  &77.62  &90.84  &86.15  &86.69  &$\textbf{69.60}$  &$\textbf{68.04}$  &73.83  &71.10  &$\textbf{68.93}$ &76.17 \\ 
		\midrule
		Baseline                        &  R-50         &88.67 	&77.62   &41.81  &58.17   &74.58  &71.64  &79.11  &90.29  &82.13  &74.32  &54.75  &60.60  &62.57  &69.67  &60.64  &68.43 \\ 
		Baseline+DAL                    &  R-50         &88.68 	&76.55   &45.08  &66.80   &67.00  &76.76  &79.74  &90.84  &79.54  &78.45  &57.71  &62.27  &69.05  &$\textbf{73.14}$  &60.11  &71.44 \\ 
        Baseline+DAL                    &  R-101         &88.61 	&79.69	&46.27	&70.37 	&65.89 	&76.10	&78.53	&90.84	&79.98	&78.41	&58.71	&62.02	&69.23	&71.32	&60.65 &71.78\\
		$\text{S}^2\text{A-Net}$\cite{han2020align}      &  R-50         &89.11  &82.84   &48.37  &71.11   &78.11  &78.39  &87.25  &90.83  &84.90  &85.64  &60.36  &62.60  &65.26  &69.13  &57.94  &74.12 \\ 
		$\text{S}^2\text{A-Net}$+DAL                     &  R-50         &89.69  &83.11  	 &$\textbf{55.03}$  &71.00   &$\textbf{78.30}$  &$\textbf{81.90}$  &$\textbf{88.46}$  &90.89  &84.97  &$\textbf{87.46}$  &64.41  &65.65  &$\textbf{76.86}$  &72.09  &64.35  &$\textbf{76.95}$  \\ 
		\bottomrule
	\end{tabular}}
\caption{  Performance evaluation of OBB task on DOTA dataset. R-101 denotes ResNet-101(likewise for R-50), and H-104 stands for Hourglass-104.}
\label{table4}
\end{table*}           
    
\begin{table}[t]
	\scriptsize
	\centering
	\setlength{\tabcolsep}{2.3mm}{
	\begin{tabular}{c|ccc|c}
		\toprule
		Methods & Backbone & Size & NA & mAP \\
		\midrule
		\multicolumn{1}{l}{\textit{Two-stage:}}&\multicolumn{4}{l}{}  \\
		\midrule
		$\text{R}^2\text{CNN}$\cite{jiang2017r2cnn}        & ResNet101   &800$\times$800     & 21   & 73.07 \\
		RC1\&RC2\cite{lb2017high}         &  VGG16      &  -                &  -   &  75.70 \\
		RRPN\cite{ma2018arbitrary}        &  ResNet101  & 800$\times$800    &  54  & 79.08  \\
		$\text{R}^2\text{PN}$\cite{zhang2018toward}        &  VGG16      &  -                &   24 &  79.60 \\
		RoI Trans. \cite{ding2019learning}&  ResNet101  & 512$\times$800    &   5  &   86.20\\
		Gliding Vertex\cite{xu2020gliding}&  ResNet101  & 512$\times$800    &   5  &  88.20 \\
		\midrule
		\multicolumn{1}{l}{\textit{Single-stage:}}&\multicolumn{4}{l}{}  \\
		\midrule
		RRD\cite{liao2018rotation}        &  VGG16      & 384$\times$384    & 13   &  84.30 \\
		$\text{R}^3\text{Det}$\cite{yang2019r3det}         &  ResNet101  & 800$\times$800    &   21 &  89.26 \\
		R-RetinaNet\cite{lin2017focal}                       &  ResNet101  & 800$\times$800    &  121 &  89.18 \\
		Baseline                          &  ResNet50   & 416$\times$416    & 3    &  80.81 \\
		Baseline+DAL                      &  ResNet50   & 416$\times$416    &  3   &  88.60 \\
		Baseline+DAL                      &  ResNet101   & 416$\times$416    & 3    &  88.95 \\
		Baseline+DAL                      &  ResNet101  & 800$\times$800    & 3    &   $\textbf{89.77}$\\
		\bottomrule
	\end{tabular}}
	\caption{Comparisons with state-of-the-art detectors on HRSC2016. NA denotes the number of preset anchor at each location of feature map.}
	\label{table5}
\end{table}
\begin{table}[t]

	\centering	
	\scriptsize
	
	\setlength{\tabcolsep}{2.2mm}{
		\begin{tabular}{c|cc|cc}
			\toprule
			Methods & car  & airplane & $\text{AP}_{50}$ & $\text{AP}_{75}$ \\ 
			\midrule
			FR-O\cite{xia2018dota}                 & 86.87 & 89.86 & 88.36 & 47.08\\ 
			RoI Transformer \cite{ding2019learning}& 87.99 & 89.90 & 88.95 & 50.54    \\
			Baseline                               & 84.64 &  $\textbf{90.51}$ & 87.57 & 39.15    \\ 
			Baseline+DAL                           & $\textbf{89.25}$ & 90.49 &  $\textbf{89.87}$ & $\textbf{74.30}$  \\ 
			\bottomrule
	\end{tabular}}
\caption{Detection results on UCAS-AOD dataset.}
	\label{table6}
\end{table}
\begin{table}[t]
	
	\centering	
	
	\scriptsize
	\setlength{\tabcolsep}{4.0mm}{
		\begin{tabular}{c|ccc}
			\toprule
			Methods & P  & R & F \\ 
			\midrule
			CTPN\cite{tian2016detecting}                                  & 74.2          & 51.6        & 60.9\\ 
			Seglink\cite{shi2017detecting}                               & 73.1          & 76.8        & 75.0   \\ 
			RRPN\cite{ma2018arbitrary}            & 82.2          & 73.2        & 77.4   \\ 
			SCRDet\cite{yang2019scrdet}                               & 81.3          & 78.9        & 80.1   \\ 
			RRD\cite{liao2018rotation}            & 85.6          & 79.0        & 82.2   \\ 
			DB\cite{liao2020real}                & $\textbf{91.8}$         & $\textbf{83.2}$        & $\textbf{87.3}$  \\ 
			\midrule
			Baseline                              &  77.2         & 77.8   & 77.5   \\ 
			Baseline+DAL                          &  83.7         & 79.5   & 81.5   \\ 
			$\text{Baseline+DAL}^{*}$                          &  84.4         & 80.5   & 82.4   \\ 
			\bottomrule
	\end{tabular}}
\caption{ Comparisons of different methods on the ICDAR 2015. P, R, F indicate recall, precision and F-measure respectively. * means multi-scale training and testing.}
\label{table7}
\end{table}

\begin{table}[t]
	\scriptsize
	\centering	
	\setlength{\tabcolsep}{3mm}{
		\begin{tabular}{ccc}
			\toprule
			Dataset               & Backbone  & mAP/F \\ 
			\midrule
			\multirow{2}{*}{ICDAR 2013} & RetinaNet &77.2\\ 
			&RetinaNet+DAL&$\textbf{81.3}$\\ 
			\midrule
			\multirow{2}{*}{NWPU VHR-10}     & RetinaNet&86.4\\ 
			&RetinaNet+DAL  &$\textbf{88.3}$ \\ 
			\midrule
			\multirow{2}{*}{VOC 2007}     & RetinaNet & 74.9\\ 
			&RetinaNet+DAL  &$\textbf{76.1}$ \\ 
			\bottomrule
	\end{tabular}}
\caption{Performance evaluation of HBB task on ICDAR 2013 and NWPU VHR-10.}
\label{table8}
\end{table}
\subsection{Implementation Details}
For the experiments, we build the baseline on RetinaNet as described above. For HRSC2016, DOTA, and UCAS-AOD, only three horizontal anchors are set with aspect ratios of \{1/2, 1, 2\}. For ICDAR, only five horizontal anchors are set with aspect ratios of \{1/5, 1/2, 1, 2, 5\}. All images are resized to 800$\times$800.  We use random flip, rotation, and HSV colour space transformation for data augmentation. 

The optimizer used for training is Adam. The initial learning rate is set to 1e-4 and is divided by 10 at each decay step. The total iterations of HRSC2016, DOTA, UCAS-AOD, and ICDAR 2015 are 20k, 30k, 15k, and 40k, respectively. We train the models on RTX 2080Ti with batch size set to 8.
\subsection{Ablation Study}
\subsubsection{Evaluation of different components}
We conduct a component-wise experiment on HRSC2016 to verify the contribution of the proposed method. The experimental results are shown in Table \ref{table1}. For variant with output IoU,  $\alpha$ is set to 0.8 for stable training, even so, detection performance still drops from 80.8\% to 78.9\%. It indicates that the output IoU is not always credible for label assignment. With the suppression of regression uncertainty, the prior space alignment and posterior feature alignment can work together effectively on label assignment, and thus performance is dramaticly improved by 4.8\% higher than the baseline. Furthermore, the model using the matching sensitivity loss function achieves mAP of 88.6\%, and the proportion of high-precision detections is significantly increased. For example, AP$_{75}$ is 9.9\% higher than the variants with uncertainty supression, which indicates that the matching degree guided loss effectively distinguishes anchors with differential localization capability, and pay more attention to high matching degree anchors to improve high-quality detection results.
\subsubsection{Hyper-parameters}
To find suitable hyperparameter settings, and explore the relationship between parameters, we conduct parameter sensitivity experiments, and the results are shown in Table \ref{table2}. In the presence of uncertainty suppression terms, as the $\alpha$ is  reduced appropriately, the influence of feature alignment increases, and the mAP increases. It indicates that the feature alignment represented by the output IoU is beneficial to select anchors with high localization capability. However, when $\alpha$ is extremely large, the performance decreases sharply. The possible reason is that most potential high quality samples are suppressed by uncertainty penalty item when the output IoU can hardly provide feedback information. In this case, weakening the uncertainty suppression ability, that is, increasing $\gamma$ helps  to alleviate this problem and make anchor selection more stable.
\begin{figure*}[t]
	\centering
	\includegraphics[width=1.8\columnwidth]{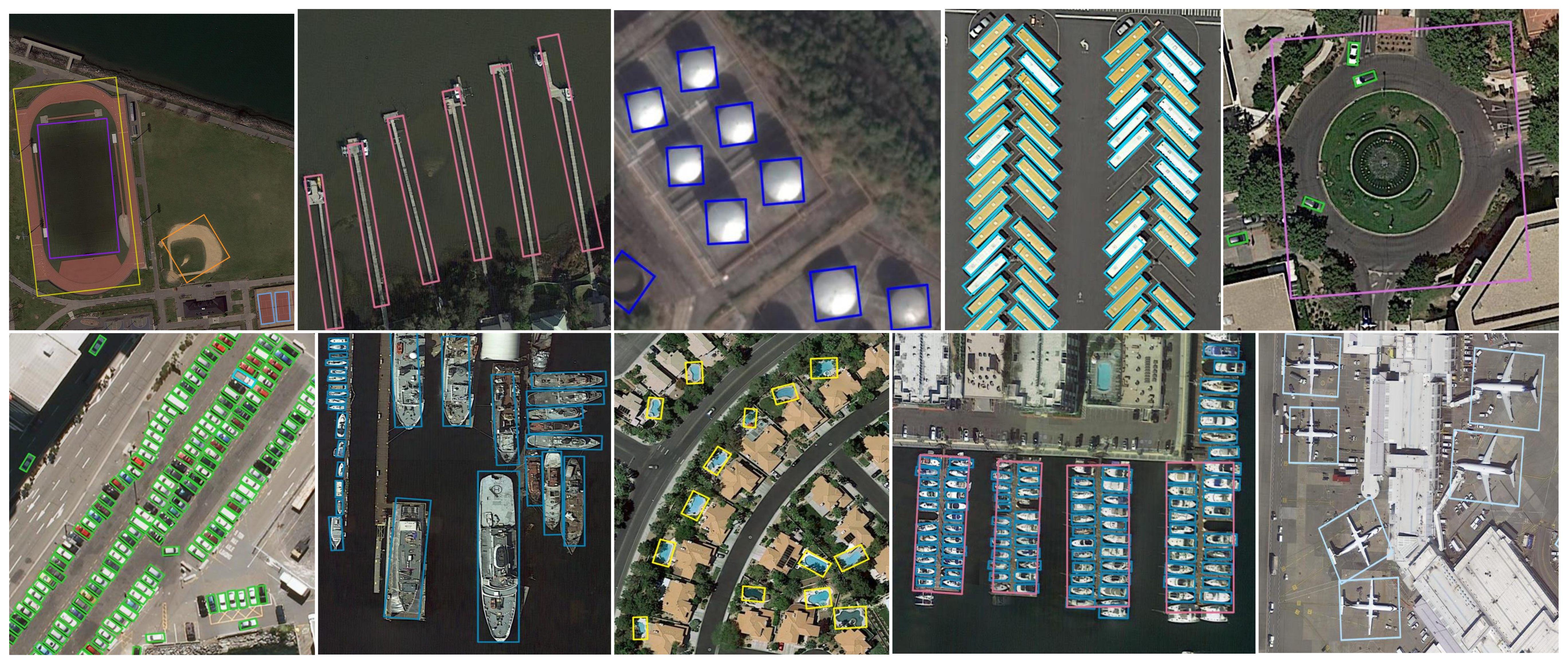} % Reduce the figure size so that it is slightly narrower than the column. Don't use precise values for figure width.This setup will avoid overfull boxes.
	\caption{ Visualization of detection results on DOTA with our method . }
	\label{fig4}
\end{figure*}

%\begin{figure}[t]
%	\centering
%	\includegraphics[width=0.8\columnwidth]{detections.png} % Reduce the figure size so that it is slightly narrower than the column. Don't use precise values for figure width.This setup will avoid overfull boxes.
%	\caption{ Visualization of detection results on HBB results . }
%	\label{fig5}
%\end{figure}
\subsection{Experiment Results} 
\subsubsection{Comparison with other sampling methods}
The experimental results are shown in Table \ref{table3}. Baseline model conducts label assignment  based on input IoU. ATSS \cite{zhang2020bridging} has achieved great success in object detection using horizontal box. When applied to rotation object detection, there is still a substantial improvement, which is 5.3\% higher than the baseline model. As for HAMBox \cite{liu2020hambox},  since the samples mined according to the output IoU are likely to be low-quality samples, too many mining samples may cause the network to fail to diverge, we only compensate one anchor for GT that do not match enough anchors. It outperforms 4.6\% than baseline. The proposed DAL method significantly increases 7.8\% compared with baseline model. Compared with the popular method ATSS, our approach considers the localization performance of the regression box, so the selected samples have more powerful localization capability, and the effect is 2.5\% higher than it, which confirms the effectiveness of our method.
\subsubsection{Results on DOTA}
We compare the proposed approach with other state-of-the-art methods. As shown in Table \ref{table4}, we achieve the mAP of 71.44\%, which outperforms the baseline model by 3\%. Integrated with DAL, even the vanilla RetinaNet can compete with many advanced methods. Besides, we also embed our approach to other models to verify its universality. S$^2$A-Net \cite{han2020align} is an advanced rotation detector that achieves the state-of-the-art performance on DOTA dataset. It can be seen that our method further improves the performance by 2.83\%, reaches the mAP of 76.95\%, achieving the best results among all compared models. Some detection results on DOTA are shown in Figure  \ref{fig4}. 
\subsubsection{Results on HRSC2016}
HRSC2016 contains lots of rotated ships with large aspect ratios and arbitrary orientation. Our method achieves the state-of-the-art performances on HRSC2016, as depicted in Table \ref{table5}. Using ResNet-101 as the backbone and the input image is resized to 800$\times$800, our method has reached the highest mAP of 89.77\%. Even if we use a lighter backbone ResNet50 and a smaller input scale of 416$\times$416, we still achieve the mAP of 88.6\%, which is comparable to many current advanced methods. 

It is worth mentioning that our method uses only three horizontal anchors in each position, but outperforms the frameworks with a large number of anchors. This shows that it is critical to effectively utilize the predefined anchors and select high-quality samples, and there is no need to preset a large number of rotated anchors. Besides, our model is a one-stage detector, and the feature map used is $P_3-P_7$. Compared with the $P_2-P_6$ for two-stage detectors, the total amount of positions that need to set anchor is less, so the inference speed is faster. With input image resized to 416$\times$416, our model reaches 34 FPS on RTX 2080 Ti GPU.

\subsubsection{Results on UCAS-AOD}
Experimental results in Table \ref{table6} show that our model achieves a further improvement of 2.3\% compared with baseline. Specifically, the detection performance of small vehicles is significantly improved, indicating that our method is also robust to small objects. Note that the DAL method dramaticly improves AP$_{75}$, which reveals that the matching degree based loss function helps to pay more attention to high-quality samples and effectively distinguish them to achieve high-quality object detection. 

\subsubsection{Results on ICDAR 2015}
To verify the generalization of our method in different scenarios, we also conduct experiments on the scene text detection dataset. The results are shown in Table \ref{table7}. Our baseline model only achieved an F-measure of 77.5\% after careful parameters selection and long-term training. The proposed DAL method improves the detection performance by 4\%, achieves an F-measure of 81.5\%. With multi-scale training and testing, it reaches 82.4\%, which was equivalent to the performance of many well-designed text detectors. However, there are a large number of long texts in ICDAR 2015 dataset, which are often mistakenly detected as several short texts. Designed for general rotation detection, DAL does not specifically consider this situation, therefore, the naive model still cannot outperform current state-of-the-art approaches for scene text detection, such as  DB \cite{liao2020real}.

\subsubsection{Experiments on Object Detection with HBB}
When localize objects with horizontal bounding box(HBB),  label assignment still suffers from the uneven discriminative feature. Although this situation is not as severe as the rotated objects, it still lays hidden dangers of unstable training. Therefore, our method is also effective in general object detection using HBB. The experimental results on ICDAR 2013 \cite{karatzas2013icdar}, NWPU VHR-10 \cite{cheng2016learning} and VOC2007 \cite{everingham2010pascal}  are shown in the Table \ref{table8}. It can be seen that DAL achieves  substantial improvements for object detection with HBB, which proves the universality of our method.
\section{Conclusion}
In this paper, we propose a dynamic anchor learning strategy to achieve high-performance arbitrary-oriented object detection. Specifically,  matching degree is constructed to comprehensively considers the spatial alignment, feature alignment ability, and regression uncertainty for label assignment. Then dynamic anchor selection and matching-sensitive loss are integrated into the training pipeline to improves the high-precision detection performance and alleviate the gap between classification and regression tasks. Extensive experiments on several datasets have confirmed the effectiveness and universality of our method.
\bibliography{refer.bib}

\begin{thebibliography}{50}
\providecommand{\natexlab}[1]{#1}
\providecommand{\url}[1]{\texttt{#1}}
\providecommand{\urlprefix}{URL }
\expandafter\ifx\csname urlstyle\endcsname\relax
  \providecommand{\doi}[1]{doi:\discretionary{}{}{}#1}\else
  \providecommand{\doi}{doi:\discretionary{}{}{}\begingroup
  \urlstyle{rm}\Url}\fi

\bibitem[{Azimi et~al.(2018)Azimi, Vig, Bahmanyar, K{\"o}rner, and
  Reinartz}]{azimi2018towards}
Azimi, S.~M.; Vig, E.; Bahmanyar, R.; K{\"o}rner, M.; and Reinartz, P. 2018.
\newblock Towards multi-class object detection in unconstrained remote sensing
  imagery.
\newblock In \emph{Asian Conference on Computer Vision}, 150--165. Springer.

\bibitem[{Cheng, Zhou, and Han(2016)}]{cheng2016learning}
Cheng, G.; Zhou, P.; and Han, J. 2016.
\newblock Learning rotation-invariant convolutional neural networks for object
  detection in VHR optical remote sensing images.
\newblock \emph{IEEE Transactions on Geoscience and Remote Sensing} 54(12):
  7405--7415.

\bibitem[{Choi et~al.(2019)Choi, Chun, Kim, and Lee}]{choi2019gaussian}
Choi, J.; Chun, D.; Kim, H.; and Lee, H.-J. 2019.
\newblock Gaussian yolov3: An accurate and fast object detector using
  localization uncertainty for autonomous driving.
\newblock In \emph{Proceedings of the IEEE International Conference on Computer
  Vision}, 502--511.

\bibitem[{Choi et~al.(2018)Choi, Lee, Lim, and Oh}]{choi2018uncertainty}
Choi, S.; Lee, K.; Lim, S.; and Oh, S. 2018.
\newblock Uncertainty-aware learning from demonstration using mixture density
  networks with sampling-free variance modeling.
\newblock In \emph{2018 IEEE International Conference on Robotics and
  Automation (ICRA)}, 6915--6922. IEEE.

\bibitem[{Dai et~al.(2016)Dai, Li, He, and Sun}]{dai2016r}
Dai, J.; Li, Y.; He, K.; and Sun, J. 2016.
\newblock R-fcn: Object detection via region-based fully convolutional
  networks.
\newblock In \emph{Advances in neural information processing systems},
  379--387.

\bibitem[{Ding et~al.(2019)Ding, Xue, Long, Xia, and Lu}]{ding2019learning}
Ding, J.; Xue, N.; Long, Y.; Xia, G.-S.; and Lu, Q. 2019.
\newblock Learning RoI transformer for oriented object detection in aerial
  images.
\newblock In \emph{Proceedings of the IEEE Conference on Computer Vision and
  Pattern Recognition}, 2849--2858.

\bibitem[{Everingham et~al.(2010)Everingham, Van~Gool, Williams, Winn, and
  Zisserman}]{everingham2010pascal}
Everingham, M.; Van~Gool, L.; Williams, C.~K.; Winn, J.; and Zisserman, A.
  2010.
\newblock The pascal visual object classes (voc) challenge.
\newblock \emph{International journal of computer vision} 88(2): 303--338.

\bibitem[{Feng, Rosenbaum, and Dietmayer(2018)}]{feng2018towards}
Feng, D.; Rosenbaum, L.; and Dietmayer, K. 2018.
\newblock Towards safe autonomous driving: Capture uncertainty in the deep
  neural network for lidar 3d vehicle detection.
\newblock In \emph{2018 21st International Conference on Intelligent
  Transportation Systems (ITSC)}, 3266--3273. IEEE.

\bibitem[{Han et~al.(2020)Han, Ding, Li, and Xia}]{han2020align}
Han, J.; Ding, J.; Li, J.; and Xia, G.-S. 2020.
\newblock Align Deep Features for Oriented Object Detection.
\newblock \emph{arXiv preprint arXiv:2008.09397} .

\bibitem[{He et~al.(2019)He, Zhu, Wang, Savvides, and Zhang}]{he2019bounding}
He, Y.; Zhu, C.; Wang, J.; Savvides, M.; and Zhang, X. 2019.
\newblock Bounding box regression with uncertainty for accurate object
  detection.
\newblock In \emph{Proceedings of the IEEE Conference on Computer Vision and
  Pattern Recognition}, 2888--2897.

\bibitem[{Jiang et~al.(2018)Jiang, Luo, Mao, Xiao, and
  Jiang}]{jiang2018acquisition}
Jiang, B.; Luo, R.; Mao, J.; Xiao, T.; and Jiang, Y. 2018.
\newblock Acquisition of localization confidence for accurate object detection.
\newblock In \emph{Proceedings of the European Conference on Computer Vision
  (ECCV)}, 784--799.

\bibitem[{Jiang et~al.(2017)Jiang, Zhu, Wang, Yang, Li, Wang, Fu, and
  Luo}]{jiang2017r2cnn}
Jiang, Y.; Zhu, X.; Wang, X.; Yang, S.; Li, W.; Wang, H.; Fu, P.; and Luo, Z.
  2017.
\newblock R2cnn: rotational region cnn for orientation robust scene text
  detection.
\newblock \emph{arXiv preprint arXiv:1706.09579} .

\bibitem[{Karatzas et~al.(2015)Karatzas, Gomez-Bigorda, Nicolaou, Ghosh,
  Bagdanov, Iwamura, Matas, Neumann, Chandrasekhar, Lu
  et~al.}]{karatzas2015icdar}
Karatzas, D.; Gomez-Bigorda, L.; Nicolaou, A.; Ghosh, S.; Bagdanov, A.;
  Iwamura, M.; Matas, J.; Neumann, L.; Chandrasekhar, V.~R.; Lu, S.; et~al.
  2015.
\newblock ICDAR 2015 competition on robust reading.
\newblock In \emph{2015 13th International Conference on Document Analysis and
  Recognition (ICDAR)}, 1156--1160. IEEE.

\bibitem[{Karatzas et~al.(2013)Karatzas, Shafait, Uchida, Iwamura, i~Bigorda,
  Mestre, Mas, Mota, Almazan, and De~Las~Heras}]{karatzas2013icdar}
Karatzas, D.; Shafait, F.; Uchida, S.; Iwamura, M.; i~Bigorda, L.~G.; Mestre,
  S.~R.; Mas, J.; Mota, D.~F.; Almazan, J.~A.; and De~Las~Heras, L.~P. 2013.
\newblock ICDAR 2013 robust reading competition.
\newblock In \emph{2013 12th International Conference on Document Analysis and
  Recognition}, 1484--1493. IEEE.

\bibitem[{Kendall and Gal(2017)}]{kendall2017uncertainties}
Kendall, A.; and Gal, Y. 2017.
\newblock What uncertainties do we need in bayesian deep learning for computer
  vision?
\newblock In \emph{Advances in neural information processing systems},
  5574--5584.

\bibitem[{Kong et~al.(2019)Kong, Sun, Liu, Jiang, and Shi}]{kong2019consistent}
Kong, T.; Sun, F.; Liu, H.; Jiang, Y.; and Shi, J. 2019.
\newblock Consistent optimization for single-shot object detection.
\newblock \emph{arXiv preprint arXiv:1901.06563} .

\bibitem[{LB et~al.(2017)}]{lb2017high}
LB, W.; et~al. 2017.
\newblock A high resolution optical satellite image dataset for ship
  recognition and some new baselines .

\bibitem[{Li, Liu, and Wang(2019)}]{li2019gradient}
Li, B.; Liu, Y.; and Wang, X. 2019.
\newblock Gradient harmonized single-stage detector.
\newblock In \emph{Proceedings of the AAAI Conference on Artificial
  Intelligence}, volume~33, 8577--8584.

\bibitem[{Li et~al.(2020)Li, Wu, Zhu, Xiong, Socher, and
  Davis}]{li2020learning}
Li, H.; Wu, Z.; Zhu, C.; Xiong, C.; Socher, R.; and Davis, L.~S. 2020.
\newblock Learning from noisy anchors for one-stage object detection.
\newblock In \emph{Proceedings of the IEEE/CVF Conference on Computer Vision
  and Pattern Recognition}, 10588--10597.

\bibitem[{Liao, Shi, and Bai(2018)}]{liao2018textboxes++}
Liao, M.; Shi, B.; and Bai, X. 2018.
\newblock Textboxes++: A single-shot oriented scene text detector.
\newblock \emph{IEEE transactions on image processing} 27(8): 3676--3690.

\bibitem[{Liao et~al.(2020)Liao, Wan, Yao, Chen, and Bai}]{liao2020real}
Liao, M.; Wan, Z.; Yao, C.; Chen, K.; and Bai, X. 2020.
\newblock Real-Time Scene Text Detection with Differentiable Binarization.
\newblock In \emph{AAAI}, 11474--11481.

\bibitem[{Liao et~al.(2018)Liao, Zhu, Shi, Xia, and Bai}]{liao2018rotation}
Liao, M.; Zhu, Z.; Shi, B.; Xia, G.-s.; and Bai, X. 2018.
\newblock Rotation-sensitive regression for oriented scene text detection.
\newblock In \emph{Proceedings of the IEEE conference on computer vision and
  pattern recognition}, 5909--5918.

\bibitem[{Lin et~al.(2017{\natexlab{a}})Lin, Doll{\'a}r, Girshick, He,
  Hariharan, and Belongie}]{lin2017feature}
Lin, T.-Y.; Doll{\'a}r, P.; Girshick, R.; He, K.; Hariharan, B.; and Belongie,
  S. 2017{\natexlab{a}}.
\newblock Feature pyramid networks for object detection.
\newblock In \emph{Proceedings of the IEEE conference on computer vision and
  pattern recognition}, 2117--2125.

\bibitem[{Lin et~al.(2017{\natexlab{b}})Lin, Goyal, Girshick, He, and
  Doll{\'a}r}]{lin2017focal}
Lin, T.-Y.; Goyal, P.; Girshick, R.; He, K.; and Doll{\'a}r, P.
  2017{\natexlab{b}}.
\newblock Focal loss for dense object detection.
\newblock In \emph{Proceedings of the IEEE international conference on computer
  vision}, 2980--2988.

\bibitem[{Liu et~al.(2016)Liu, Anguelov, Erhan, Szegedy, Reed, Fu, and
  Berg}]{liu2016ssd}
Liu, W.; Anguelov, D.; Erhan, D.; Szegedy, C.; Reed, S.; Fu, C.-Y.; and Berg,
  A.~C. 2016.
\newblock Ssd: Single shot multibox detector.
\newblock In \emph{European conference on computer vision}, 21--37. Springer.

\bibitem[{Liu, Ma, and Chen(2018)}]{liu2018arbitrary}
Liu, W.; Ma, L.; and Chen, H. 2018.
\newblock Arbitrary-oriented ship detection framework in optical remote-sensing
  images.
\newblock \emph{IEEE Geoscience and Remote Sensing Letters} 15(6): 937--941.

\bibitem[{Liu et~al.(2020)Liu, Tang, Han, Liu, Rui, and Wu}]{liu2020hambox}
Liu, Y.; Tang, X.; Han, J.; Liu, J.; Rui, D.; and Wu, X. 2020.
\newblock HAMBox: Delving Into Mining High-Quality Anchors on Face Detection.
\newblock In \emph{2020 IEEE/CVF Conference on Computer Vision and Pattern
  Recognition (CVPR)}, 13043--13051. IEEE.

\bibitem[{Liu et~al.(2017)Liu, Hu, Weng, and Yang}]{liu2017rotated}
Liu, Z.; Hu, J.; Weng, L.; and Yang, Y. 2017.
\newblock Rotated region based CNN for ship detection.
\newblock In \emph{2017 IEEE International Conference on Image Processing
  (ICIP)}, 900--904. IEEE.

\bibitem[{Ma et~al.(2018)Ma, Shao, Ye, Wang, Wang, Zheng, and
  Xue}]{ma2018arbitrary}
Ma, J.; Shao, W.; Ye, H.; Wang, L.; Wang, H.; Zheng, Y.; and Xue, X. 2018.
\newblock Arbitrary-oriented scene text detection via rotation proposals.
\newblock \emph{IEEE Transactions on Multimedia} 20(11): 3111--3122.

\bibitem[{Pan et~al.(2020)Pan, Ren, Sheng, Dong, Yuan, Guo, Ma, and
  Xu}]{pan2020dynamic}
Pan, X.; Ren, Y.; Sheng, K.; Dong, W.; Yuan, H.; Guo, X.; Ma, C.; and Xu, C.
  2020.
\newblock Dynamic Refinement Network for Oriented and Densely Packed Object
  Detection.
\newblock In \emph{Proceedings of the IEEE/CVF Conference on Computer Vision
  and Pattern Recognition}, 11207--11216.

\bibitem[{Pang et~al.(2019)Pang, Chen, Shi, Feng, Ouyang, and
  Lin}]{pang2019libra}
Pang, J.; Chen, K.; Shi, J.; Feng, H.; Ouyang, W.; and Lin, D. 2019.
\newblock Libra r-cnn: Towards balanced learning for object detection.
\newblock In \emph{Proceedings of the IEEE conference on computer vision and
  pattern recognition}, 821--830.

\bibitem[{Redmon et~al.(2016)Redmon, Divvala, Girshick, and
  Farhadi}]{redmon2016you}
Redmon, J.; Divvala, S.; Girshick, R.; and Farhadi, A. 2016.
\newblock You only look once: Unified, real-time object detection.
\newblock In \emph{Proceedings of the IEEE conference on computer vision and
  pattern recognition}, 779--788.

\bibitem[{Ren et~al.(2015)Ren, He, Girshick, and Sun}]{ren2015faster}
Ren, S.; He, K.; Girshick, R.; and Sun, J. 2015.
\newblock Faster r-cnn: Towards real-time object detection with region proposal
  networks.
\newblock In \emph{Advances in neural information processing systems}, 91--99.

\bibitem[{Shi, Bai, and Belongie(2017)}]{shi2017detecting}
Shi, B.; Bai, X.; and Belongie, S. 2017.
\newblock Detecting oriented text in natural images by linking segments.
\newblock In \emph{Proceedings of the IEEE Conference on Computer Vision and
  Pattern Recognition}, 2550--2558.

\bibitem[{Shrivastava, Gupta, and Girshick(2016)}]{shrivastava2016training}
Shrivastava, A.; Gupta, A.; and Girshick, R. 2016.
\newblock Training region-based object detectors with online hard example
  mining.
\newblock In \emph{Proceedings of the IEEE conference on computer vision and
  pattern recognition}, 761--769.

\bibitem[{Song, Liu, and Wang(2020)}]{song2020revisiting}
Song, G.; Liu, Y.; and Wang, X. 2020.
\newblock Revisiting the sibling head in object detector.
\newblock In \emph{Proceedings of the IEEE/CVF Conference on Computer Vision
  and Pattern Recognition}, 11563--11572.

\bibitem[{Tian et~al.(2016)Tian, Huang, He, He, and Qiao}]{tian2016detecting}
Tian, Z.; Huang, W.; He, T.; He, P.; and Qiao, Y. 2016.
\newblock Detecting text in natural image with connectionist text proposal
  network.
\newblock In \emph{European conference on computer vision}, 56--72. Springer.

\bibitem[{Wei et~al.(2019)Wei, Zhou, Zhang, Li, Guo, and
  Wang}]{wei2019oriented}
Wei, H.; Zhou, L.; Zhang, Y.; Li, H.; Guo, R.; and Wang, H. 2019.
\newblock Oriented objects as pairs of middle lines.
\newblock \emph{arXiv preprint arXiv:1912.10694} .

\bibitem[{Xia et~al.(2018)Xia, Bai, Ding, Zhu, Belongie, Luo, Datcu, Pelillo,
  and Zhang}]{xia2018dota}
Xia, G.-S.; Bai, X.; Ding, J.; Zhu, Z.; Belongie, S.; Luo, J.; Datcu, M.;
  Pelillo, M.; and Zhang, L. 2018.
\newblock DOTA: A large-scale dataset for object detection in aerial images.
\newblock In \emph{Proceedings of the IEEE Conference on Computer Vision and
  Pattern Recognition}, 3974--3983.

\bibitem[{Xu et~al.(2020)Xu, Fu, Wang, Wang, Chen, Xia, and
  Bai}]{xu2020gliding}
Xu, Y.; Fu, M.; Wang, Q.; Wang, Y.; Chen, K.; Xia, G.-S.; and Bai, X. 2020.
\newblock Gliding vertex on the horizontal bounding box for multi-oriented
  object detection.
\newblock \emph{IEEE Transactions on Pattern Analysis and Machine Intelligence}
  .

\bibitem[{Yang et~al.(2019{\natexlab{a}})Yang, Liu, Yan, Li, Zhang, and
  Yu}]{yang2019r3det}
Yang, X.; Liu, Q.; Yan, J.; Li, A.; Zhang, Z.; and Yu, G. 2019{\natexlab{a}}.
\newblock R3det: Refined single-stage detector with feature refinement for
  rotating object.
\newblock \emph{arXiv preprint arXiv:1908.05612} .

\bibitem[{Yang et~al.(2018)Yang, Sun, Fu, Yang, Sun, Yan, and
  Guo}]{yang2018automatic}
Yang, X.; Sun, H.; Fu, K.; Yang, J.; Sun, X.; Yan, M.; and Guo, Z. 2018.
\newblock Automatic ship detection in remote sensing images from google earth
  of complex scenes based on multiscale rotation dense feature pyramid
  networks.
\newblock \emph{Remote Sensing} 10(1): 132.

\bibitem[{Yang and Yan(2020)}]{yang2020arbitrary}
Yang, X.; and Yan, J. 2020.
\newblock Arbitrary-Oriented Object Detection with Circular Smooth Label.
\newblock \emph{arXiv preprint arXiv:2003.05597} .

\bibitem[{Yang et~al.(2019{\natexlab{b}})Yang, Yang, Yan, Zhang, Zhang, Guo,
  Sun, and Fu}]{yang2019scrdet}
Yang, X.; Yang, J.; Yan, J.; Zhang, Y.; Zhang, T.; Guo, Z.; Sun, X.; and Fu, K.
  2019{\natexlab{b}}.
\newblock Scrdet: Towards more robust detection for small, cluttered and
  rotated objects.
\newblock In \emph{Proceedings of the IEEE International Conference on Computer
  Vision}, 8232--8241.

\bibitem[{Zhang, Lu, and Zhang(2019)}]{zhang2019cad}
Zhang, G.; Lu, S.; and Zhang, W. 2019.
\newblock Cad-net: A context-aware detection network for objects in remote
  sensing imagery.
\newblock \emph{IEEE Transactions on Geoscience and Remote Sensing} 57(12):
  10015--10024.

\bibitem[{Zhang et~al.(2020{\natexlab{a}})Zhang, Chang, Ma, Wang, and
  Chen}]{zhang2020dynamic}
Zhang, H.; Chang, H.; Ma, B.; Wang, N.; and Chen, X. 2020{\natexlab{a}}.
\newblock Dynamic R-CNN: Towards High Quality Object Detection via Dynamic
  Training.
\newblock \emph{arXiv preprint arXiv:2004.06002} .

\bibitem[{Zhang et~al.(2020{\natexlab{b}})Zhang, Chi, Yao, Lei, and
  Li}]{zhang2020bridging}
Zhang, S.; Chi, C.; Yao, Y.; Lei, Z.; and Li, S.~Z. 2020{\natexlab{b}}.
\newblock Bridging the gap between anchor-based and anchor-free detection via
  adaptive training sample selection.
\newblock In \emph{Proceedings of the IEEE/CVF Conference on Computer Vision
  and Pattern Recognition}, 9759--9768.

\bibitem[{Zhang et~al.(2019)Zhang, Wan, Liu, Ji, and Ye}]{zhang2019freeanchor}
Zhang, X.; Wan, F.; Liu, C.; Ji, R.; and Ye, Q. 2019.
\newblock Freeanchor: Learning to match anchors for visual object detection.
\newblock In \emph{Advances in Neural Information Processing Systems},
  147--155.

\bibitem[{Zhang et~al.(2018)Zhang, Guo, Zhu, and Yu}]{zhang2018toward}
Zhang, Z.; Guo, W.; Zhu, S.; and Yu, W. 2018.
\newblock Toward arbitrary-oriented ship detection with rotated region proposal
  and discrimination networks.
\newblock \emph{IEEE Geoscience and Remote Sensing Letters} 15(11): 1745--1749.

\bibitem[{Zhu et~al.(2015)Zhu, Chen, Dai, Fu, Ye, and
  Jiao}]{zhu2015orientation}
Zhu, H.; Chen, X.; Dai, W.; Fu, K.; Ye, Q.; and Jiao, J. 2015.
\newblock Orientation robust object detection in aerial images using deep
  convolutional neural network.
\newblock In \emph{2015 IEEE International Conference on Image Processing
  (ICIP)}, 3735--3739. IEEE.

\end{thebibliography}
\end{document}